\documentclass[10pt,journal,compsoc]{IEEEtran}

\ifCLASSOPTIONcompsoc
  \usepackage[nocompress]{cite}
\else
  \usepackage{cite}
\fi

\ifCLASSOPTIONcompsoc
 \usepackage[caption=false,font=footnotesize,labelfont=sf,textfont=sf]{subfig}
\else
 \usepackage[caption=false,font=footnotesize]{subfig}
\fi

\usepackage{url}
\usepackage{subfig}
\usepackage{comment}
\usepackage{algorithm}
\usepackage{algorithmic}
\usepackage{amssymb}
\usepackage{amsthm}
\usepackage{bm}
\usepackage{bbm}
\usepackage{booktabs}
\usepackage{amsmath}

\newtheorem*{rep@theorem}{\rep@title}
\newcommand{\newreptheorem}[2]{%
\newenvironment{rep#1}[1]{%
 \def\rep@title{#2 \ref{##1}}%
 \begin{rep@theorem}}%
 {\end{rep@theorem}}}
\newreptheorem{theorem}{Theorem}
\newreptheorem{lemma}{Lemma}
\newreptheorem{claim}{Claim}

\usepackage{mathtools}
\usepackage{color}
\usepackage{xcolor}
\usepackage[noabbrev]{cleveref}

\usepackage{etoolbox}
\usepackage{multirow}
\makeatletter
\patchcmd{\@makecaption}
  {\scshape}
  {}
  {}
  {}
\makeatother
\usepackage{color, colortbl}
\usepackage{mathtools}
\usepackage{esvect}
\definecolor{Back}{gray}{0.8}

\begin{document}
\title{
  Spatiotemporal-Augmented Graph Neural Networks for Human Mobility Simulation
}

\author{
  Yu~Wang,
	Tongya~Zheng*,
	Shunyu~Liu,
    Zunlei~Feng,
	Kaixuan~Chen,
	Yunzhi~Hao,
	Mingli~Song

	\IEEEcompsocitemizethanks{
	\IEEEcompsocthanksitem Yu Wang, Shunyu Liu, Kaixuan Chen, Yunzhi Hao are with the College of Computer Science, Zhejiang university, Hangzhou, China.\protect
 	\IEEEcompsocthanksitem Tongya Zheng is with Big Graph Center, School of Computer and Computing Science, Hangzhou~City~University, Hangzhou, China.\protect
	\IEEEcompsocthanksitem Zunlei Feng is with the College of Software Technology, Zhejiang University, Hangzhou, China.\protect
	\IEEEcompsocthanksitem Mingli Song is with the State Key Laboratory of Blockchain and Security, Zhejiang University, Hangzhou, China.\protect  
  \IEEEcompsocthanksitem *Corresponding author: doujiang\_zheng@163.com.
	}
}

\IEEEtitleabstractindextext{
\begin{abstract}
Human mobility patterns have shown significant applications in policy-decision scenarios and economic behavior researches. 
The human mobility simulation task aims to generate human mobility trajectories given a small set of trajectory data, which have aroused much concern due to the scarcity and sparsity of human mobility data. 
Existing methods mostly rely on the static relationships of locations, while largely neglect the dynamic spatiotemporal effects of locations. 
On the one hand, spatiotemporal correspondences of visit distributions reveal the spatial proximity and the functionality similarity of locations. 
On the other hand, the varying durations in different locations hinder the iterative generation process of the mobility trajectory. 
Therefore, we propose a novel framework to model the dynamic spatiotemporal effects of locations, namely \textbf{S}patio\textbf{T}emporal-\textbf{A}ugmented g\textbf{R}aph neural networks~(STAR). 
The STAR framework designs various spatiotemporal graphs to capture the spatiotemporal correspondences and builds a novel dwell branch to simulate the varying durations in locations, which is finally optimized in an adversarial manner.
The comprehensive experiments over four real datasets for the human mobility simulation have verified the superiority of STAR to \emph{state-of-the-art} methods. 
Our code is available at \url{https://github.com/Star607/STAR-TKDE}.
\end{abstract}

}

\maketitle
\IEEEdisplaynontitleabstractindextext

\IEEEpeerreviewmaketitle

\IEEEraisesectionheading{\section{Introduction}\label{sec:introduction}}

\IEEEPARstart{H}{uman} mobility patterns have attracted widespread attention~\cite{gonzalez2008understanding,althoff2017large,jia2020population} to investigate when and where people's activities happen, and aroused multi-disciplinary applications such as urban planning~\cite{xu2021emergence}, pollution abatement~\cite{bohm2022gross}, and epidemic prevention~\cite{venkatramanan2021forecasting}.
For example, reliable human mobility models can inform policy-making of crucial projects, such as planning for expanding urban land~\cite{gao2020mapping} and making effective controlling strategies of gross polluters~\cite{bohm2022gross}.
Specifically, the human mobility patterns have been investigated to deal with the COVID-19 pandemic, appropriately informing the reopening policies~\cite{chang2021mobility} and allocating the limited medical resources~\cite{venkatramanan2021forecasting}.

\begin{figure}[t]
    \centering
    \includegraphics[width=\linewidth]{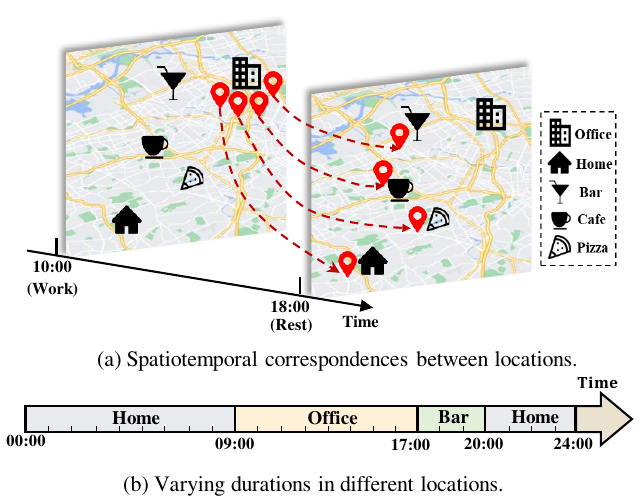}
    \caption{An illustration example. (a) depicts that various locations (HOME, Bar, Cafe, Pizza) get busy when people get off work from the office, revealing the spatiotemporal correspondences among locations.
    (b) depicts the successive visits of locations in an individual trajectory, resulting in varying durations in different locations.}
    \label{fig:spatiotemporal}
\end{figure}

However, it is difficult to obtain human mobility trajectories recording the spatiotemporal sequences of human mobility locations due to privacy concerns and commercial limitations.
The human mobility simulation task aims to generate massive artificial human mobility trajectories with high fidelity given a small set of real-world human mobility data, which can largely alleviate the scarcity and sparsity of existing data.
Such controllable generations of human mobility data can perform counterfactual inference to measure the treatment effects of different strategies~\cite{chang2021mobility} and investigate the emergence patterns of urban growth when increasing human mobility scales~\cite{alessandretti2020scales,xu2021emergence}.

Capturing the regularities of human mobility behaviors lies at the core of human mobility simulation. 
Previous model-based methods~\cite{gambs2012next,yin2017generative} build various conditional transition matrices of locations and maximize the log-likelihood by fitting the real-world data.
Despite the simplicity and interpretability, these methods rely on large-scale mobility trajectory data and fine-grained labels of location categories, which limits their practical utility.
Recently, deep learning-based methods~\cite{feng2018deepmove,feng2020learning,gao2022contextual} introduce the static relationships of locations to generate human mobility sequences in a global view.
DeepMove~\cite{feng2018deepmove} captures the sequential dependency within a single trajectory with the neural attention layer;
CGE~\cite{gao2022contextual} obtains the global relationships of locations by the proposed static relationship matrix of locations;
MoveSim~\cite{feng2020learning} takes a step further to introduce the structure prior of locations with an attention layer.
However, these methods focus on the static relationships of locations, while the dynamic spatiotemporal effects of locations are largely underexplored.

As shown in Figure~\ref{fig:spatiotemporal}, the spatiotemporal effects of locations can be observed from two aspects: the spatiotemporal correspondences and the varying durations.
On the one hand, Figure~\ref{fig:spatiotemporal}~(a) depicts that various locations (Home, Bar, Cafe, Pizza) get busy when people get off work. 
It indicates spatiotemporal correspondences between these locations, which can reflect both the spatial proximity and the functionality similarity.
On the other hand, Figure~\ref{fig:spatiotemporal}~(b) depicts the varying durations in different locations of an illustrative trajectory, which describes the spatiotemporal continuity of human mobility behaviors.
Existing methods generate the simulation trajectories without considering the varying durations, where locations with short dwell time will undoubtedly get neglected in the optimization goals.

Therefore, in this paper, we propose a novel \textbf{S}patio\textbf{T}emporal-\textbf{A}ugmented g\textbf{R}raph neural networks (STAR) to model the spatiotemporal effects of locations in a generator-discriminator paradigm.
Firstly, we construct various kinds of spatiotemporal graphs to capture the spatiotemporal correspondences of locations and obtain the location embeddings with the multi-channel embedding module.
Secondly, we build a dual-branch decision generator module to capture the varying durations in different locations, where the exploration branch accounts for the diverse transitions of locations and the dwell branch accounts for the staying patterns of locations.
After generating a complete human trajectory iteratively, the proposed STAR is optimized by the policy gradient strategy with rewards from the policy discriminator module, playing a min-max game with the discriminator~\cite{goodfellow2020generative}.
We have conducted comprehensive experiments on four real datasets for the human mobility simulation task.
Results on various real-world datasets validate the superiority of our proposed STAR framework to the \textit{state-of-the-art} methods.
In summary, our contributions can be summarized as follows:
\begin{itemize}
  \item We innovatively build the spatiotemporal graphs of locations to model the dynamic spatiotemporal effects among locations for the human mobility simulation task.
  \item {A novel framework STAR is proposed to handle spatiotemporal correspondences and varying durations with the multi-channel embedding module and the dual-branch decision generator module, respectively.}
  \item {Extensive experiments on various datasets demonstrate that our proposed STAR consistently outperforms the \textit{state-of-the-art} baselines in human mobility simulation with high fidelity. The ablation studies further reveal the working mechanisms of the spatiotemporal graphs and the dwell branch.}
\end{itemize}

\section{Related Works}
In this section, we briefly review the most-related literatures along the 
following lines of fields: (1) human mobility simulation and (2) graph
neural networks.

\subsection{Human Mobility Simulation}
The human mobility simulation task aims to generate artificial mobility trajectories with realistic mobility patterns given a small set of human mobility data \cite{karamshuk2011human, hess2015data, shin2020user, wang2024spatiotemporal, wang2024cola}. 
The generated artificial trajectories must reproduce a set of spatial and temporal mobility patterns, such as the distribution of characteristic distances and the predictability of human whereabouts.
Temporal patterns usually include the number and sequence of visited locations together with the time and duration of the visits, which involves balancing an individual's routine and sporadic out-of-routine mobility patterns.
And spatial patterns include the preference for short distances\cite{gonzalez2008understanding, pappalardo2013understanding}, the tendency to split into returnees and explorers\cite{pappalardo2015returners}, and the fact of visiting multiple sites for constant times for individuals\cite{alessandretti2018evidence}. 

In the early stages, Markov-based models dominate the human mobility simulation task.
For example, the first-order MC~\cite{song2003evaluating} defines the state as the accessed location and assumes that the next location solely depends on the current location, thus constructing a transition matrix to capture the first-order transition probability among locations.
HMM~\cite{krumm2004locadio} is established with discrete emission probability and optimized by the Baum-Welch algorithm.
IO-HMM~\cite{yin2017generative} further extends HMM by introducing more annotation information, which also improves interpretability.
However, Markov-based models are limited in capturing long-term dependencies and incorporating individual preferences.

To make up for the deficiency of Markov-based models, a large body of mechanistic methods have emerged, which can reproduce basic temporal, spatial, and social patterns of human mobility \cite{barbosa2018human, pappalardo2017data, karamshuk2011human, hess2015data, wang2019urban}.
For example, Exploration and Preferential Return (EPR) model\cite{song2010modelling} enables an agent to select a new location that has never been visited before based on a random walk process utilizing a power-law jump-size distribution, or return to a previously visited location based on its frequency.
Then several studies enhanced the EPR model by incorporating increasingly elaborate spatial or social mechanisms \cite{pappalardo2015returners, barbosa2015effect, alessandretti2018evidence, toole2015coupling, cornacchia2021stsepr}.
EPR and its extensions primarily focus on capturing the spatial patterns of human mobility, neglecting to capture the temporal mechanisms. 
TimeGeo \cite{jiang2016timegeo} and DITRAS \cite{pappalardo2017data} improve the temporal mechanism by integrating a data-driven model into an EPR-like model to capture both routine and out-of-routine circadian preferences.
Despite the interpretability of mechanistic methods, their realism is limited by the simplicity of the implemented mechanisms.

The limitations mentioned above can be addressed by deep learning generative paradigms such as recurrent neural networks (RNNs) and generative adversarial networks (GANs), which can learn the distribution of data by capturing complex and non-linear relationships in data and generate mobility trajectories from the same distribution. 
RNN-based models prefer to maximize the prediction likelihood of the next location likelihood~\cite{bengio2015scheduled,yu2017seqgan}, resulting in ignoring the long-term influence and the so-called \textit{exposure bias}.
SeqGAN~\cite{yu2017seqgan} is the pioneering work of sequence generation based on GAN~\cite{goodfellow2020generative}.
MoveSim~\cite{feng2020learning} extends SeqGAN by introducing the location structure as a prior for human mobility simulation.
ActSTD~\cite{yuan2022activity} improves the dynamic modeling of individual trajectories by the neural ordinary equation.
SAND~\cite{yuan2023learning} combines generative adversarial imitation learning with Maslow's need theory to simulate human activities.
It is worth noting that human mobility simulation is distinct from time series prediction based on mobility~\cite{kapoor2020examining, bao2022covid}, where the former aims to generate a sequence of discrete locations while the latter focuses on predicting a real number.
Furthermore, different from human mobility prediction~\cite{nweke2018deep, jaouedi2020prediction} and location recommendation~\cite{pan2022location}, human mobility simulation emphasizes producing trajectories that reflect the characteristics of real-world data while protecting user privacy, rather than testing the model's ability to recover the real data.  
Despite deep learning approaches proposed for human mobility prediction, simulating daily mobility has been underexplored.

\begin{figure*}[!t]
    \centering
    \includegraphics[width=\textwidth]{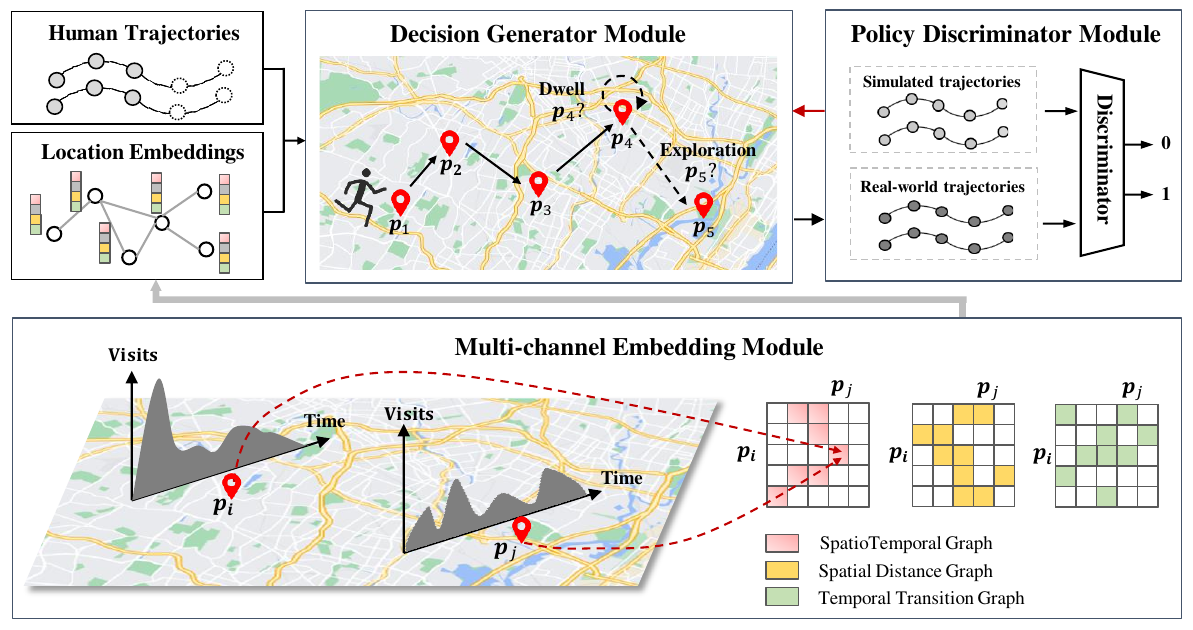}
    \caption{
    The overall framework of STAR.
    Firstly, given an observed human trajectory, the multi-channel embedding module generates location embeddings based on the proposed multi-channel spatiotemporal graphs.
    Secondly, the decision generator module predicts the future trajectory by balancing the exploration branch which is prone to another location and the dwell branch which decides whether to stay at the previous location.  
    Finally, STAR is optimized in an adversarial manner by the policy discriminator module to alleviate the \emph{exposure bias} of the maximum likelihood manner.}
    \label{fig:star}
\end{figure*}

\subsection{Graph Neural Networks}
Recently, the rapid development of deep learning~\cite{jwang_paper1,jwang_paper3} has inspired the researches on Graph Neural Networks~(GNNs)~\cite{deffe2016gcn,kipf2016semi,hamilton2017graphsage,velivckovic2017gat}.
Due to the significant progress of GNNs, they have been applied in various fields, where urban computing is most relevant to the human mobility simulation task in this paper.
Urban computing aims to understand the urban patterns and dynamics from different application domains where the big data explodes, such as transportation, environment, security, etc~\cite{zheng2014urban}. 
Due to the spatio-temporal characteristics of some typical urban data, such as traffic network flow~\cite{pan2020spatio, wang2021exploring, liu2022traffic, zeng2021modeling}, crowd flow~\cite{gong2020online}, environmental monitoring data, etc., some previous works combine graph neural networks with various temporal learning networks to capture the dynamics in the spatial and temporal dimensions~\cite{wang2020deep}. 
The hybrid neural network architecture is collectively referred to as spatio-temporal graph neural network (STGNN).

The basic STGNN framework for predictive learning is composed of three modules: Data Processing Module (DPM) which constructs the spatio-temporal graph from raw data, Spatio-Temporal Graph Learning Module (STGLM) which extracts hidden spatio-temporal dependencies within complex social systems and Task-Aware Prediction Module (TPM) which maps the spatio-temporal hidden representation from STGLM into the space of downstream prediction tasks. 
As the most crucial part in STGNN, STGLM combines spatial learning networks such as spectral GCNs, spatial GCNs or GATs, and temporal learning networks such as RNNs, temporal convolutional networks (TCNs) or temporal self-attention networks (TSANs) organically through a certain spatio-temporal fusion neural architecture. 

Therefore, almost all current researches focus on the design of the neural architectures in STGLM and there are many frontier methods to improve the learning of spatio-temporal dependencies. 
For example, THINK~\cite{agarwal2022think} and DMGCRN~\cite{qin2021dmgcrn} perform the hyperbolic graph neural network on the Poincare ball to directly capture multi-scale spatial dependencies. 
ASTGCN~\cite{guo2019attention} employ a typical three-branch architecture for multi-granularity temporal learning, where the data undergoes calculations of multiple GCNs and Attention networks from the three branches will be finally fused using the learnable weight matrix. 
STSGCN~\cite{song2020spatial} fuses spatio-temporal dependencies by constructing the spatio-temporal synchronous graph.
Based on STSGNN, STFGNN~\cite{li2021spatial} introduces both topology-based graph and similarity-based graph to construct a spatio-temporal synchronous graph, making the spatio-temporal synchronous graph more informative. 
S2TAT~\cite{wang2022synchronous} proposes a spatio-temporal synchronous transformer framework to enhance the learning capability with attention mechanisms. 

However, there are two main differences between our work and the existing researches on STGNN~\cite{guo2019attention, song2020spatial, li2021spatial, qin2021dmgcrn, agarwal2022think, wang2022synchronous, du2022disentangled}. 
First, the topological structure of the spatio-temporal graph (e.g. road network) in previous studies is fixed, while ours is self-constructed and closely related to the order of locations visited by individuals. 
Second, the existing methods based on STGNN are almost used for predictive learning tasks in urban computing, but our approach focuses on the simulation task which emphasizes the effective capture of the overall patterns rather than the accurate prediction of a single entity.

\section{Problem Statement}

A human mobility trajectory can be defined as a spatiotemporal sequence $s=[\tau_1, \tau_2, \cdots, \tau_L]$.
The $l$-th visit record $\tau_l$ represents a tuple $(p_l, t_l)$, where $p_l$ records the location ID and $t_l$ records the visiting timestamp.
The human mobility simulation problem is thus defined as follows.

\textsc{DEFINITION 1. (Human Mobility Simulation)}.
\emph{Given a real-world mobility trajectory dataset $\mathcal{S} = \{s_1, s_2, \cdots, s_m\}$, our goal is to learn to simulate human mobility behaviors in order to generate an artificial trajectory $\hat{s}=\left[\hat{\tau}_1, \hat{\tau}_2, \dots, \hat{\tau}_L\right]$ with fidelity and utility.}

Directly maximizing the likelihood of the sequence generation model would undoubtedly cause the \textit{exposure bias}~\cite{bengio2015scheduled,yu2017seqgan} towards the training data, resulting in poor generalization abilities.
Therefore, it is useful to advance the human mobility simulation under the framework of SeqGAN~\cite{goodfellow2020generative,yu2017seqgan}, which coordinates the optimization of the generator (sequence generation model) and a discriminator in a min-max game, written as
\begin{equation}
  \min _{\mathbf{\theta}} \max _{\mathbf{\phi}}
  \mathbb{E}_{\mathbf{x} \sim \pi} 
  {\left[\log D_\mathbf{\phi}(\mathbf{x})\right]}
  +
  \mathbb{E}_{\mathbf{x} \sim \pi_d}
  \left[\log \left(1-D_\mathbf{\phi}(\mathbf{x})\right)\right],
\end{equation}
where $\mathbf{x}$ is the sampled mobility trajectory, $\mathbb{E}_\pi$ represents the expected reward of the sequences under the policy $\pi$, and $\mathbb{\pi}_d$ samples $\mathbf{x}$ from the ground-truth trajectory data $\mathcal{S}$.

\section{Methods}
In this section, we propose \textbf{S}patio\textbf{T}emporal-\textbf{A}ugmented g\textbf{R}aph neural networks~(STAR) to simulate human mobility, whose framework is illustrated in Figure~\ref{fig:star}.
In this framework, we firstly develop the multi-channel embedding module to capture the spatiotemporal correspondences among locations based on the proposed multi-channel location graphs.
Secondly, we design the decision generator module involving exploration and dwell branches to learn the varying durations in different locations from trajectories with highly repetitive patterns. 
Thirdly, we optimize the training process using the policy discriminator module, which provides the rewards for the generator at each step based on the policy gradient technique.

\subsection{Multi-channel Embedding Module} 
It's fundamental to acquire high-quality representations of locations with ample spatiotemporal semantics for effective human mobility simulation.
Considering the non-grid structure of locations, we leverage GNNs to model the spatiotemporal dependencies among the locations.
However, relying solely on static spatial distance or temporal transition, or both, is insufficient to capture the intricate spatiotemporal relationships.
Therefore, to go beyond the geographical effects, we first build a Spatial Distance Graph (SDG) and Temporal Transition Graph (TTG) based on the observed human mobility trajectories, which are constructed from the spatial and temporal perspectives, respectively.
Secondly, we propose a SpatioTemporal Graph (STG) by measuring the Wasserstein distance~\cite{panaretos2019statistical} of location visit distributions to capture the spatiotemporal effects, which combines the spatial and temporal dynamics of locations simultaneously.
Finally, we fuse the representations of locations from these three graphs to obtain comprehensive semantics.
In this part, we first describe the details of constructing location graphs, followed by an explanation of how location embeddings are generated.

\subsubsection{Location Graphs}

Let $\mathcal{P}=\{p_0,p_1,\cdots,p_N\}$ be the set of locations.
The spatiotemporal correspondences among locations $p_i, p_j \in \mathcal{P}$ can be well modeled by their correspondence scores $\epsilon(p_i, p_j)$ with a correspondence function $\epsilon(\cdot, \cdot)$, resulting in a location-location adjacency graph $G=(\mathcal{P}, \mathcal{E})$.
The edge set $\mathcal{E}$ of the graph contains pairwise correspondences between locations.
For example, if the function $\epsilon(\cdot, \cdot)$ measures the geographical proximity of locations, the constructed graph $G$ represents a $K$-nearest neighbor spatial graph of all locations.
Based on the spatial graph, our model will generate human mobility trajectories in consideration of geographical effects.
Next, we describe the construction of Spatial Distance Graph, Temporal Transition Graph and SpatioTemporal Graph respectively.

\textbf{Spatial Distance Graph.}
The proposed SDG is constructed based on the spatial proximity of a location pair $(p_i, p_j)$, which indicates the spatial cooperation effects in human trajectories.
Let $w_{ij}$ be the spatial proximity score of $\epsilon(p_i,p_j)$, we have to build the Spatial Distance Graph out of the Cartesian score set of  $\mathcal{P}\times\mathcal{P}$, written as:
\begin{equation}
    G_{\mathrm{SDG}} = 
    \begin{cases}
    w_{ij}, & w_{ij} \in \text{top}_k(\epsilon(p_i, \cdot)),\\
    0, & \text{Otherwise},
    \end{cases}
    \label{eq:sdg}
\end{equation}
which remains the top-$k$ neighbors of all locations.
The spatial proximity function is usually implemented as Euclidean distance for simplicity.

\textbf{Temporal Transition Graph.}
The SDG can reveal the spatial proximity of locations, but it ignores mobility patterns of human trajectories.
Consequently, we further propose the TTG to encode mobility patterns based on the partially observed human trajectories.
Let $s=[\cdots, \tau_l, \tau_{l+1}, \cdots]$ be a human trajectory and $\tau_l=(p_i, t_l), \tau_{l+1}=(p_j, t_{l+1})$. 
We record the proximity score between $p_i$ and $p_j$ as $\epsilon(p_i,p_j)=1$ and obtain the summed scores by $w_{ij} = \sum_{(pi,p_j) \in \mathcal{S}} \epsilon(p_i, p_j)$.
Then the TTG can be formulated as:
\begin{equation}
    G_{\mathrm{TTG}} = 
    \begin{cases}
    w_{ij}, & w_{ij} > 0,\\
    0, & \text{Otherwise}.
    \end{cases}
    \label{eq:ttg}
\end{equation}
The obtained TTG can provide detailed descriptions of mobility patterns from the temporal transition perspective.

\textbf{SpatioTemporal Graph.}
The SDG and TTG describe the relationships between locations from the spatial and temporal perspectives, respectively, ignoring the other aspect to some extent.
Therefore, we further characterize the functionality of locations with the STG by their visit distribution over time, which can describe the spatial proximity by spatial cooperation effects and the temporal proximity by the similarity in visit distribution.
For instance, cafes often open from morning to evening, while bars keep open till midnight.
The visit distribution $F_{p_i}$ of a location $p_i$ is its normalized visit counts in the observed records $\mathcal{S}_{p_i}$ by discretizing the timestamps into $T$ time slots: $F_{p_i}(t) = \frac{1}{\vert \mathcal{S}_{p_i} \vert }\sum_{t_k \in \mathcal{S}_{p_i}} \mathbbm{1}{[t_k=t]}$.
The obtained visit distribution $F_{p_i}$ can well describe the location $p_i$ from both spatial and the temporal aspects.
On the one hand, neighbor locations often share similar visit distributions based on geographical effects, which have been widely adopted in Location-based Services.
On the other hand, locations far apart can also share similar visit distributions as they provide comparable functionalities like eating or drinking.
Thus, the Wasserstein Distance~\cite{panaretos2019statistical, lan2022dstagnn} is employed here to quantify the differences among the visit distributions $F_{\mathcal{P}}$ of locations to differentiate their individual functionalities, defined as follows:  
\begin{equation}
  \begin{gathered}
  d(p_i, p_j)=\inf _{\pi \in \prod[F_{p_i}, F_{p_j}]} \int_x \int_y \pi(x, y) |x-y| \mathrm{d} x \mathrm{d} y, \\ 
  \text { s.t. } \int \pi(x, y) \mathrm{d} y= F_{p_i}(x),
  \int \pi(x, y) \mathrm{d} x= F_{p_j}(y),
  \end{gathered}
  \label{eq:distance}
\end{equation}
where $\pi(x,y)$ is a joint distribution of $F_{p_i}$ and $F_{p_j}$.
The proximity score can be simply defined as $\epsilon(p_i, p_j) = 1 - d(p_i, p_j) \in [0, 1]$.
Let $w_{ij}=\epsilon(p_i,p_J)$, the STG can be formulated as:
\begin{equation}
    G_{\mathrm{STG}} = 
    \begin{cases}
    w_{ij}, &  w_{ij} \in \text{top}_k(\epsilon(p_i, \cdot)),\\
    0, & \text{Otherwise}.
    \end{cases}
    \label{eq:stg}
\end{equation}
The proposed STG can capture spatiotemporal dynamics of locations in a distribution way.

Overall, the above multi-channel location graphs are denoted by $\mathcal{G}=\{G_{\mathrm{SDG}}, G_{\mathrm{TTG}}, G_{\mathrm{STG}}\}$.
The connection edges of a location pair $(p_i,p_j) \in G$ are naturally weighted with their proximity scores, indicating the spatiotemporal correspondences given by different measurements.
However, these scores are not aligned with the human mobility simulation task, which may cause the inconsistent problems between the spatiotemporal graphs and the human mobility simulation task.
To bridge the gap between the graphs and the task, we alternatively propose a \emph{Vanilla} version based on the \emph{Weighted} version, written as:
\begin{equation}
  \begin{split}
    \hat{\epsilon}(p_i, p_j)=
    \begin{cases}
    1, & (p_i, p_j) \in G,\\
    0, & \text{Otherwise}.\\
    \end{cases}
  \end{split}
  \label{eq:graph-binary}
\end{equation}
The \emph{Vanilla} graph regards all neighbors as the same, benefiting the model optimization without the need for computed weights of the multi-channel graphs.

\subsubsection{Location Embeddings}
After establishing the above graphs, location embeddings can be obtained by the following procedure.

Firstly, we implement the features of the location set $\mathcal{P}$ at the input layer by a learnable embedding matrix with end-to-end optimization, following the common training paradigm of human mobility simulation.

Secondly, to adaptively aggregate the useful information in terms of the graph heterogeneity, we adopt the attention layer of graphs~\cite{velivckovic2017gat} to perform multi-channel attention over all kinds of spatiotemporal graphs, namely $\mathcal{G}=\{G_{\mathrm{SDG}}, G_{\mathrm{TTG}}, G_{\mathrm{STG}}\}$.
Given location embeddings $\mathbf{H}^{l-1}_G$ from the previous $l-1$-th layer (initialized embeddings at the input layer), our spatiotemporal attention layer computes the node embeddings in a multi-head manner by:
\begin{equation}
  \mathbf{h}^{l}_i = \Big\|_{k=1}^{K} {\text{ReLU}} \left( \sum_{j \in \mathcal{N}_i} \alpha^k_{ij} W^k \mathbf{h}_j^{l-1} \right), 
  \label{eq:multi-head}
\end{equation}
where $\alpha_{ij}^k$ is the $k$-th attention scores of the edge $(p_i, p_j)$, $\mathcal{N}_i$ provides the neighbors of location $p_i$, $K$ is the number of attention heads, $\|$ is the concatenation operation, $W^k$ is the transformation matrix of $k$-th attention head, and ReLU is a non-linear activation function.

For simplicity, the $l$-th layer location embeddings are summed up over all graphs: $\mathbf{H}^l = \sum_{G \in \mathcal{G}} \mathbf{H}^l_G$.
The final embeddings of locations are denoted by $\mathbf{H}_{\mathcal{P}}$.

\subsection{Decision Generator Module}
The multi-channel embedding module provides location embeddings via our dedicated spatiotemporal graphs, enhancing the representation quality used in the following decision generator module.
The decision generator module autoregressively predicts the next location based on the observed partial mobility trajectory.
To cope with the varying durations of locations as shown in Figure~\ref{fig:spatiotemporal}~(b), we develop the dwell branch besides the common exploration branch in GANs, which also reduces the optimization difficulty to some extent.  
Since the policy discriminator will not give classification results until the trajectory sequence is finished, we adopt the Monte-Carlo strategy to iteratively predict the next location until the required length.

\subsubsection{Exploration Branch}
Suppose that a partial mobility trajectory $s_{1:l} = \{(p_1, t_1), (p_2, t_2), \cdots, (p_l, t_l)\}$ is observed, 
the exploration branch aims to predict the next location $\hat{p}_{l+1}$ that maximizes the decision reward given by the policy discriminator module, where $t_{l+1}$ is assumed uniformly spaced with $t_l$ for simplicity.
Given a bunch of location embeddings $\mathbf{H}_{\mathcal{P}}$ from the multi-channel embedding module, we firstly fetch the corresponding input embeddings of the trajectory, denoted by $\mathbf{h}_s = \{\mathbf{h}_{p_0}, \mathbf{h}_{p_1}, \cdots, \mathbf{h}_{p_l}\}$.
To facilitate the sequence modeling and training efficiency, we adopt the Gated Recurrent Unit~\cite{cho2014properties} as our sequence model, written as:
\begin{equation}
    \mathbf{h}_{l}, z_{l} = \mathrm{GRU}(\mathbf{h}_{p_l}, z_{l-1}),
    \label{eq:gru}
\end{equation}
where the hidden state $z_{l-1}$ is initialized as zero vectors in the first step.
The obtained sequence representation $\mathbf{h}_l$ is fed into a linear layer to predict the probability of the next locations by:
\begin{equation}
  \hat{p}_{l+1}  =\mathrm{softmax}(\mathbf{h}_l \times \mathbf{W}_p +\mathbf{b}_p).
  \label{eq:next_pred}
\end{equation}
Then the outputs of the exploration branch $\hat{p}_{l+1}$ together with the outputs of the dwell branch determine the next location iteratively until the sequence is completed.

\subsubsection{Dwell Branch}
Motivated by the varying durations of locations in mobility trajectories, we specially design a dwell branch to predict the probability of staying at the previous location, which can inform the model to \emph{dwell} or \emph{explore}. 
Intuitively, the duration of a specific location is determined by the spatiotemporal context and the cumulative duration of a trajectory, where the former can be well captured by the exploration branch and the latter lacks specific modeling.
For a highly repetitive trajectory dataset, the optimization goal will be overwhelmed by the repetitive locations, leading to the ignorance of diverse behaviors.
Therefore, it is necessary to build the dwell branch to predict whether to stay at the previous location to alleviate the learning difficulties of the exploration branch.
Taking the sequence representation $\mathbf{h}_l$ for the dwell branch classification is straightforward.
However, it ignores the saturation of human mobility behaviors~\cite{gonzalez2008understanding} that the durations of locations are usually upper-bounded to some extent.
We thus combine the sigmoid function with an exponentially decaying coefficient to reduce the effects of a long-duration location, written as:
\begin{equation}
  \hat{y}_{l+1}^{d}=\mathrm{sigmoid}(\mathbf{h}_l \mathbf{W}_d +\mathbf{b}_d) \cdot \exp (-{\beta} \cdot \mathrm{C}(s_{1:l}, p_l)),
\end{equation}
where the hyper-parameter $\beta$ adjusts the decaying rate and $\mathrm{C}(s_{1:l}, p_l)=\sum_{p_k \in s_{1:l}} \mathbbm{1}[p_k = p_l]$ counts the frequency of $p_l$ in the observed trajectory.
$\beta$ is set to 1 by default.

Finally, the decision generator module balances the location predictions $\hat{p}_{l+1}$ of the exploration branch and the dwell predictions $\hat{y}_{l+1}^d$ by sampling from the output distributions, written as:
\begin{equation}
  \begin{split}
    p_{l+1} =
    \begin{cases}
      p_l, & l>1 \; \text{and} \; \Psi(\hat{y}^d_{l+1}) = 1, \\
      \Psi(\hat{p}_{l+1}), & \text{Otherwise}, \\
    \end{cases}
  \end{split}
  \label{eq:group-evolution}
\end{equation}
where $\Psi(.)$ samples a location by polynomial sampling from a given probability distribution. 

\subsection{Policy Discriminator Module}
As discussed in~\cite{bengio2015scheduled,yu2017seqgan}, maximizing the likelihood of sequence models would suffer from the \textit{exposure bias} of the training data, thus generalizing poorly to the testing data.
Recently, variants of GANs~\cite{yu2017seqgan,feng2020learning} have shown their successful applications in sequence generation tasks by back-propagating the policy gradients into the generator.
Therefore, we follow the generator-discriminator paradigm to perform human mobility simulation in an adversarial manner.

By generating and sampling iteratively in the decision generator module, a set of generated trajectories $\mathcal{T}_{G}$ is fed into the policy discriminator module $D_{\phi}$.
Since a powerful discriminator might hinder the optimization of the generator $G_{\theta}$, we build a simple yet efficient discriminator $D_{\phi}$ that contains an embedding matrix of locations and classifies the trajectory sequence as real data or fake data based on a GRU layer, written as:
\begin{equation}
  \mathcal{L}_D=\mathbb{E}_{\mathbf{x} \in \mathcal{T}_R} \log D_\phi(\mathbf{x})+\mathbb{E}_{(\mathbf{x}) \in \mathcal{T}_G} \log \left(1-D_\phi(\mathbf{x})\right),
\end{equation}
where $\mathcal{T}_R$ and $\mathcal{T}_G$ are real and generated trajectories, respectively. 
The decision generator module $G_{\theta}$ makes stepwise predictions and 
receives the policy gradients from the discriminator $D_{\phi}$ by unfolding the classification reward recursively following the REINFORCE algorithm~\cite{williams1992simple}, written as:
\begin{equation}
  \nabla_\theta=\nabla_\theta \mathbb{E}_{P_\theta(\mathbf{x})}[R(\mathbf{x})]=\mathbb{E}_{P_\theta(\mathbf{x})}\left[R(\mathbf{x}) \nabla_\theta \log P_\theta(\mathbf{x})\right],
\end{equation}
where $P_\theta$ denotes the generated distribution by the decision generator $G_{\theta}$, $\mathbf{x}$ is the generated mobility trajectory, and the reward $R(\mathbf{x})$ is the classification loss from the discriminator $D_{\phi}$. 
The generator $G_\theta$ can be optimized in this way.

\section{Experiment}
\label{sec:experiment}
In this section, we conduct experiments for the human mobility simulation task on four real-world datasets to evaluate the performance of our proposed STAR framework. 
We first briefly introduce the four datasets, baseline methods, and evaluation metrics.
Then, we compare STAR with the \textit{state-of-the-art} baselines for the human mobility simulation task and present the experiment results.
Furthermore, we analyze the impact of different modules and hyperparameters on model performance. 
Finally, we present the geographical visualization results of our proposed STAR method and the baseline methods.

We aim to answer the following key research questions:
\begin{itemize}
  \item \textbf{RQ1}: How does STAR perform compared with other \textit{state-of-the-art} methods for the human mobility simulation task?
  \item \textbf{RQ2}: How do different modules (different kinds of spatiotemporal graphs and the dwell branch) affect STAR?
  \item \textbf{RQ3}: How do hyper-parameter settings (the depth of layer and the number of attention heads) influence the performance of STAR?
  \item \textbf{RQ4}: How do we qualitatively assess the quality of human trajectories simulated by different methods with regard to the real trajectories?

\end{itemize}

\begin{table}[t]
  \caption{The statistics of four datasets.}
  \centering
  \resizebox{1.0\linewidth}{!}{
  \begin{tabular}{lcccc}
    \toprule
    Dataset  & Timespan & $\vert$Users$\vert$ & $\vert$Locations$\vert$ & $\vert$Visits$\vert$  \\
    \midrule
    NYC   & Apr. 2012 - Feb. 2013 &1,083  &38,333  &227,428  \\
    TKY  & Apr. 2012 - Feb. 2013 &2,293  &61,858  &573,703  \\
    Moscow   & Apr. 2012 - Sep. 2013  &10,464  &88,036  &806,196 \\
    Singapore  & Apr. 2012 - Sep. 2013  &8,784  &45,525  &397,873   \\
    \bottomrule
  \end{tabular}
  }
  \label{tab:data}
\end{table}

\begin{table*}[ht]
  \caption{
  Performance of the proposed STAR framework and baselines in terms of JSD for human mobility trajectory simulation. 
  All experimental results are conducted over five trials for a fair comparison.
  A lower JSD value indicates a better performance. 
  \textbf{Bold} and \underline{underline} means the best and the second-best results.
  ``*" implies statistical significance for $p < 0.05$ under paired t-test.
  }
  \centering
  \resizebox{1.0\textwidth}{!}{
  \begin{tabular}{l|cccccc|cccccc}
    \toprule
    & \multicolumn{6}{c}{ \textbf{NYC}} & \multicolumn{6}{c}{ \textbf{ TKY}} \\
    \hline               				
    \textbf{Metrics(JSD)} & Distance  & Radius  & Duration  & DailyLoc  & G-rank & I-rank & Distance  & Radius  & Duration  & DailyLoc  & G-rank & I-rank \\
    \hline
    Markov\cite{gambs2012next} & 0.1253 & \underline{0.3711} & 0.0037 & 0.2686 & 0.0966 & 0.0673  & 0.1373 & \underline{0.4249} & 0.0079 & 0.2627 & 0.0420 & 0.0624 \\
    IO-HMM\cite{yin2017generative} & 0.4364 & 0.6170 & 0.0025 & 0.0834 & 0.4053 & \textbf{0.0435}  & 0.3598 & 0.5514 & 0.0044 & 0.1885 & 0.2325 & 0.0498  \\
    LSTM\cite{lstm1997hoch} & \underline{0.1240} & 0.3830 & 0.0030 & 0.0862 & \underline{0.0723} & 0.0564 & \underline{0.1249} & 0.4341 & 0.0019 & \underline{0.0426} & \underline{0.0342} & 0.0638 \\
    DeepMove\cite{feng2018deepmove} & 0.4292 & 0.6347 & 0.0050 & 0.3113 & 0.1091 & 0.0521  & 0.3518 & 0.5791 & \underline{0.0010} & 0.0480 & 0.2232 & 0.0522 \\
    GAN\cite{goodfellow2020generative} & 0.4407 & 0.6272 & 0.0046 & 0.0813 & 0.2420 & \underline{0.0482}  & 0.3633 & 0.5615 & 0.0044 & 0.1871 & 0.1679 & \textbf{0.0457}  \\
    SeqGAN\cite{yu2017seqgan} & 0.1737 & 0.4536 & 0.0028 & \underline{0.0626} & 0.0822 & 0.0563 & 0.1304 & 0.4344 & 0.0011 & 0.0483 & 0.0428 & 0.0571  \\
    MoveSim\cite{feng2020learning} & 0.2932 & 0.5471 & \underline{0.0021} & 0.3903 & 0.1872 & 0.0574  & 0.2225 & 0.4368 & 0.0054 & 0.4651 & 0.1494 & 0.0510 \\
    CGE\cite{gao2022contextual} & 0.4098 & 0.5880 & 0.0598 & 0.3189 & 0.3170 & 0.0499  & 0.3913 & 0.6310 & 0.0762 & 0.3984 & 0.1891 & \underline{0.0473} \\
    VOLUNTEER\cite{long2023practical} & 0.2498 & 0.6811 & 0.6652 & 0.6931 & 0.1160 & 0.0778  & 0.1804 & 0.5570 & 0.6595 & 0.6931 & 0.0605 & 0.0692 \\
    \hline
    \specialrule{0em}{1pt}{1pt}
    STAR~(Ours) & \textbf{0.1134}* & \textbf{0.3615}* & \textbf{0.0012}* & \textbf{0.0597} & \textbf{0.0378}* & 0.0550  & \textbf{0.1206}* & \textbf{0.4198}* & \textbf{0.0004}* & \textbf{0.0340}* &\textbf{0.0307}*  & 0.0650 \\
    \specialrule{0em}{1pt}{1pt}
    \toprule
    & \multicolumn{6}{c}{ \textbf{Moscow}} & \multicolumn{6}{c}{ \textbf{Singapore}} \\
    \hline
    \textbf{Metrics(JSD)}  & Distance  & Radius  & Duration  & DailyLoc  & G-rank & I-rank & Distance  & Radius  & Duration  & DailyLoc  & G-rank & I-rank \\
    \hline
    Markov\cite{gambs2012next} & 0.0090 & 0.0589 & 0.0135 & 0.1690 & 0.1231 & 0.0666 & 0.0093 & 0.0389 & 0.0060 & 0.1425 & 0.2517 & 0.0783 \\
    IO-HMM\cite{yin2017generative} & 0.1048 & 0.0778 & 0.0096 & 0.2492 & 0.2158 & 0.0724 & 0.0509 & 0.0447 & 0.0041 & 0.1595 & 0.3569 & 0.0577 \\
    LSTM\cite{lstm1997hoch} & 0.0117 & 0.0194 & 0.0075 & 0.1373 & 0.0358 & \underline{0.0618} & \underline{0.0084} & \underline{0.0111} & 0.0113 & 0.1190 & 0.1186 & 0.0680 \\
    DeepMove\cite{feng2018deepmove} & 0.1048 & 0.0854 & 0.0050 & 0.1283 & 0.2552 & 0.0647 & 0.0413 & 0.0338 & \underline{0.0012} & \underline{0.0469} & 0.1410 & \underline{0.0474} \\
    GAN\cite{goodfellow2020generative} & 0.1059 & 0.0746 & 0.0096 & 0.2476 & 0.2158 & 0.0714 & 0.0504 & 0.0433 & 0.0043 & 0.1600 & 0.3325 & 0.0577 \\
    SeqGAN\cite{yu2017seqgan} & 0.0222 & 0.0314 & \underline{0.0043} & \underline{0.0615} & \underline{0.0268} & 0.0631 & 0.0145 & 0.0149 & 0.0037 & 0.0924 & \textbf{0.0269} & 0.0520 \\
    MoveSim\cite{feng2020learning} & 0.0828 & 0.0598 & 0.0114 & 0.2739 & 0.1404 & 0.0644 & 0.0222 & 0.0125 & 0.0110 & 0.3473 & 0.1718 & 0.0552 \\
    CGE\cite{gao2022contextual} & 0.0647 & 0.0981 & 0.0097 & 0.2319 & 0.1968 & 0.0698 & 0.0906 & 0.0494 & 0.0802 & 0.4196 & 0.3286 & 0.0559 \\ 
    VOLUNTEER\cite{long2023practical} & \underline{0.0087} & \underline{0.0186} & 0.6352 & 0.6795 & 0.0843 & 0.0625 & 0.0119 & 0.0112 & 0.6619 & 0.6931 & 0.1137 & 0.0666 \\
    \hline
    \specialrule{0em}{1pt}{1pt}
    STAR~(Ours) & \textbf{0.0071} & \textbf{0.0165}* & \textbf{0.0022}* & \textbf{0.0516} & \textbf{0.0196}* & \textbf{0.0518}* & \textbf{0.0080} & \textbf{0.0108} & \textbf{0.0009} & \textbf{0.0330}* & \underline{0.0283} & \textbf{0.0418}* \\
    \specialrule{0em}{1pt}{1pt}
    \bottomrule

  \end{tabular}
  }
  \label{tab:baseline}
\end{table*}

\subsection{Datasets}
To ensure reproducibility and facilitate fair comparisons with previous works, we evaluate the human mobility of four cities (i.e., New York, Tokyo, Moscow and Singapore) extracted from the publicly available dataset Foursquare in line with previous studies~\cite{gao2022contextual, feng2017poi2vec}.
The timespan and the numbers of users, locations and visit records in each dataset are shown in Table~\ref{tab:data}.
The wide range of the timespan and the large-scale visits can sufficiently compare the STAR framework against baseline methods.
In experiments, we use the original raw datasets that only contain the GPS coordinates of each location and user check-in records, and pre-process them following the protocol of the human mobility simulation task. 

Specifically, we split the whole dataset into three parts: a training set for training the generative model, a validation set for finding the best parameters of models and a testing set for the final evaluation of various metrics. 
The partition of the four datasets is set as 7:1:2. 
Besides, we set the basic time slot as an hour of the day for the convenience and universality of modeling. 
Finally, in order to ensure the effectiveness and accuracy of modeling, we only remain the trajectories with visit records greater than eight each day.

\subsection{Baseline Methods and Evaluation Metrics}
We compare the performance of the STAR framework with the following \textit{state-of-the-art} baseline methods.
\begin{itemize}
  \item 
  \textbf{Markov}~\cite{gambs2012next} is a well-known probability method describing the state transitions, which treats the locations as states and calculates the transition probability of locations.
  \item 
  \textbf{IO-HMM}~\cite{yin2017generative} fits the probability model with annotated user activities as its latent states and then generates human trajectories based on the hidden Markov model.
  \item 
  \textbf{LSTM}~\cite{lstm1997hoch} improves the classical RNN by the dedicated memory cell and the forget gate to enhance the long-term dependency modeling.
  \item 
  \textbf{DeepMove}~\cite{feng2018deepmove} employs an attentional recurrent network to acquire the knowledge of sequential transitions and periodic movement patterns from lengthy and sparse trajectories.
  \item 
  \textbf{GAN}~\cite{goodfellow2020generative} uses two LSTMs as the generator and the discriminator in our settings, respectively.
  \item
  \textbf{SeqGAN}~\cite{yu2017seqgan} solves the problem of generating discrete sequence by leveraging the policy gradient technique in conjunction with GAN.
  \item 
  \textbf{MoveSim}~\cite{feng2020learning} simulates mobility trajectory by augmenting the prior knowledge of the generator and regularizing the periodicity of sequences.
  \item 
  \textbf{CGE}~\cite{gao2022contextual} generates location sequences based on a newly constructed static graph from historical visit records.
  \item 
  \textbf{VOLUNTEER}~\cite{long2023practical} proposes a user VAE and a trajectory VAE to capture mobility patterns from both group and individual views.
\end{itemize}

Following the common practice in previous works~\cite{feng2020learning, ouyang2018non}, we adopt six metrics to evaluate the quality of generated data by comparing the distributions of important mobility patterns between the simulated trajectories and the real trajectories from different perspectives.
\begin{itemize}
  \item \textbf{Distance}: The moving distance among locations in individuals' trajectories is a metric from the spatial perspective. 
  \item \textbf{Radius}: Radius of gyration is the root mean square distance of all locations from the central one, which represents the spatial range of individual daily movement.
  \item \textbf{Duration}: Dwell duration among locations in mobility trajectories is a metric from the temporal perspective. 
  \item \textbf{DailyLoc}: Daily visited locations are calculated as the number of visited locations per day for each user. 
  \item \textbf{G-rank}: The number of visits per location is calculated as the visiting frequency of the top-100 locations. 
  \item \textbf{I-rank}: It is an individual version of G-rank.
\end{itemize}
Specifically, we use Jensen-Shannon divergence (JSD)~\cite{fuglede2004jensen} to measure the discrepancy of the distributions between the generated data and real-world data. 
The JSD metric is defined as follows:
\begin{equation}
  \operatorname{JSD}(p \| q)=H((p+q) / 2)-\frac{1}{2}(H(p)+H(q))
\end{equation}
where $p$ and $q$ are two distributions for comparison, and $H$ is the Shannon entropy.  
A lower JSD denotes a closer match to the statistical characteristics and thus indicates a better generation result.

\subsection{Experimental Settings}
\label{sec: exp_set}
  (1) Baseline methods with adversarial learning process (i.e. GAN, SeqGAN, MoveSim) use a hidden dimension of 32 for generator and a hidden dimension of 64 for discriminator. 
  In addition, they are pre-trained with 100 epochs and adversarially trained with 50 epochs.
  (2) All other baseline methods use a hidden dimension of 32 and are trained with 50 epochs.
  (3) All methods use a batch size of 32.
  The other specific hyperparameters of the baselines follow the settings reported in their respective papers.
  For STAR, the number of GAT layers is searched over \{1, 2, 3\}, and the number of attention heads is searched over \{1, 2, 4, 8\}.

\subsection{RQ1: Performance Comparison}

Table~\ref{tab:baseline} presents the performance in retaining the data fidelity of our framework and the eight competitive baselines on four real-world datasets at different scales.
The results reveal the following discoveries:
\begin{itemize}
  \item \textbf{Our framework steadily achieves the best performance.} 
  STAR achieves the best performance on all datasets with ranking first on twenty-one metrics and ranking second on one metric over twenty-four metrics of the four datasets.   
  For the twenty-one ranking first metrics, compared with the best baseline, our method reduces the JSD up to 60\%. 
  For the other three metrics, namely \textit{I-rank} on the NYC and TKY dataset, \textit{G-rank} on the Singapore dataset, our framework also obtains competitive performance with the best baseline.  
  \item \textbf{Model-based methods are limited in simulating human mobility.}
  Markov performs worse on the time-dependent metrics (i.e., \textit{Duration} and \textit{Dailyloc}) but better on the distance-based metrics (i.e., \textit{Distance} and \textit{Radius}), because it obtains the next location based on the distribution of historical transition probabilities, which aligns with our intuition that individuals are more likely to visit nearby areas.
  The performance of IO-HMM is also unsatisfactory, because its modeling relies on extensive manual labeling, which places higher demands on the quality of data. 
  For example, lack of records at home interferes with the annotation of \textsl{Home} label, which degrades the predictive performance of IO-HMM.
  In addition, the sparsity in data introduces errors when labeling dwell time.
  \item \textbf{Deep learning methods for mobility prediction task perform poorly on hunam mobility simulation task.}
  LSTM and DeepMove are all trained with the short-term goal (i.e. next location prediction) and thus do not perform particularly well on the human mobility simulation task. 
  Unlike human mobility simulation, which emphasizes producing results that reflect various mobility patterns in real trajectories, human mobility prediction highlights the recovery of the next location in real-data and lacks the learning of the global patterns.
  \item \textbf{Generative networks fail to generate realistic human trajectories without trajectory pre-training.}
  GAN performs the worst across almost all metrics, indicating that it is difficult to capture the hidden patterns of human mobility when learning with noisy and inaccurate raw data.
  SeqGAN is pre-trained by the task based on human mobility modeling, so it yields much better results than GAN.
  MoveSim introduces the urban structure modeling component especially for locations, so it achieves the second-best result on \textit{Duration} metric of NYC dataset.
  \item \textbf{It is essential to model dynamic spatiotemporal dependencies among locations.}
  Although CGE leverages graph structure for data augmentation, its graph is static and the node embeddings generated by Word2vec cannot be learned and updated in the process of mobility simulation, so it only learns better on several metrics but performs poorly overall.
  VOLUNTEER achieves the second-best results on the \textit{Distance} and \textit{Radius} metrics due to its specific modeling for workplace and residence distribution from the group view.
\end{itemize}

\subsection{RQ2: Ablation Study of STAR}

\begin{figure*}[t]
  \includegraphics[width=\textwidth]{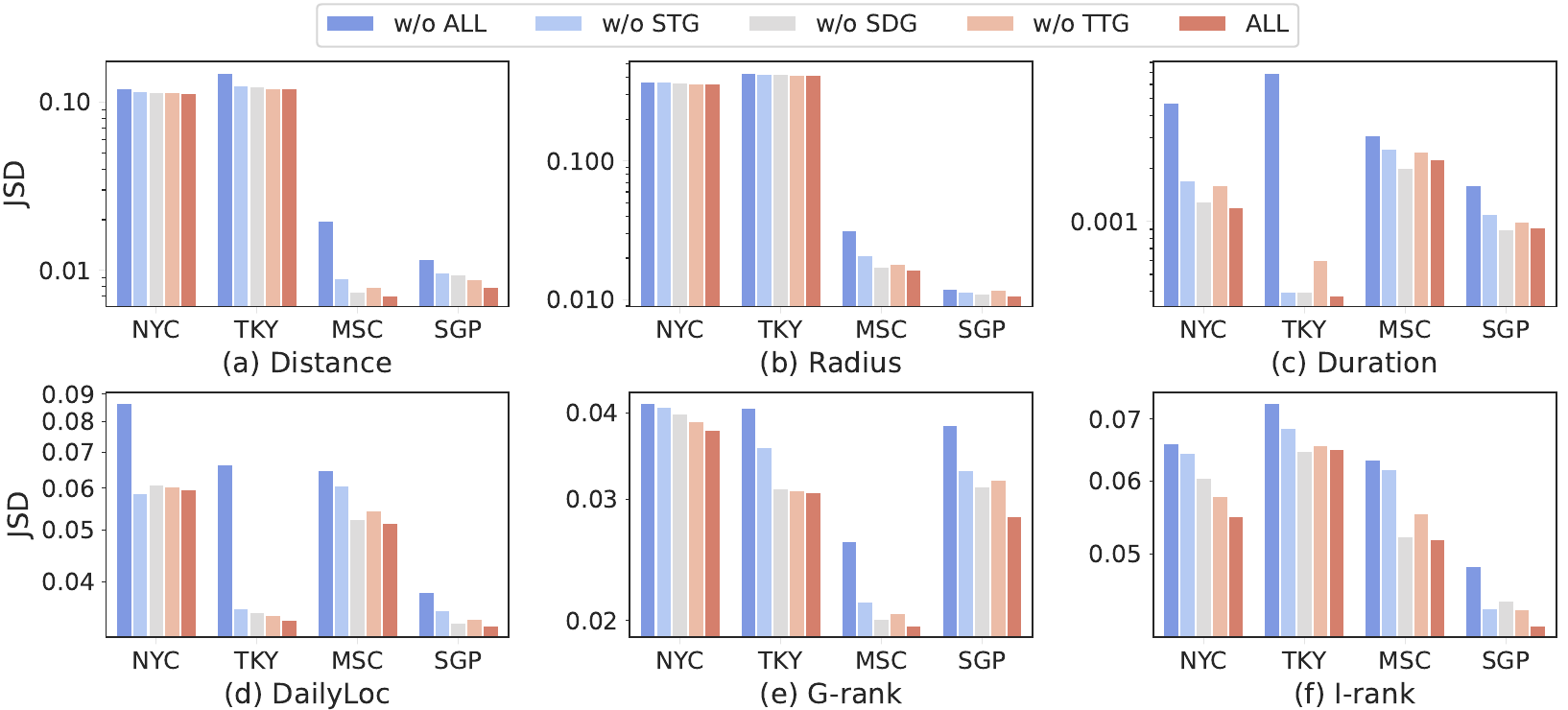}
  \caption{Ablation study on the channels of graphs.
  All experimental results are conducted over five trials for a fair comparison.
  STG, SDG and TTG represent SpatioTemporal Graph, Spatial Distance Graph and Temporal Transition Graph respectively. 
  MSC and SGP are short for Moscow and Singapore. 
  A lower JSD value indicates a better performance. }
  \label{fig:gid_res}
\end{figure*}

\begin{table}
\caption{Ablation study on edge type of GNNs in STAR in terms of JSD. 
  All experimental results are conducted over five trials for a fair comparison.
MSC and SGP are short for Moscow and Singapore. A lower JSD value indicates a better performance.}
\centering
\resizebox{1.0\linewidth}{!}{
  \begin{tabular}{llcccccc}
  \toprule
  Dataset & Method & Distance & Radius & Duration & DailyLoc & G-rank & I-rank \\
  \midrule
  \multirow{3}{*}{NYC} & Weighted & 0.1150 & 0.3672 & 0.0014 & 0.0613 & 0.0387 & 0.0584 \\
  & Vanilla & 0.1134 & 0.3615 & 0.0012 & 0.0597 & 0.0378 & 0.0550 \\
  & \% Improv. & 1.43\% & 1.56\% & 14.29\% & 2.68\% & 2.33\% & 5.89\% \\
  \midrule
  \multirow{3}{*}{TKY} & Weighted & 0.1245 & 0.4256 & 0.0005 & 0.0351 & 0.0326 & 0.0693 \\
  & Vanilla & 0.1206 & 0.4198 & 0.0004 & 0.0340 & 0.0307 & 0.0650 \\
  & \% Improv. & 3.13\% & 1.36\% & 24.00\% & 3.25\% & 5.89\% & 6.20\% \\
  \midrule
  \multirow{3}{*}{MSC} & Weighted & 0.0103 & 0.0234 & 0.0031 & 0.0524 & 0.0206 & 0.0555 \\
  & Vanilla & 0.0071 & 0.0165 & 0.0022 & 0.0516 & 0.0196 & 0.0518 \\
  & \% Improv. & 31.07\% & 29.66\% & 27.74\% & 1.60\% & 4.66\% & 6.63\% \\
  \midrule
  \multirow{3}{*}{SGP} & Weighted & 0.0112 & 0.0144 & 0.0011 & 0.0322 & 0.0310 & 0.0427 \\
  & Vanilla & 0.0080 & 0.0108 & 0.0009 & 0.0330 & 0.0283 & 0.0418 \\
  & \% Improv. & 28.57\% & 25.14\% & 16.36\% & - & 8.58\% & 2.15\% \\ 
  \bottomrule
  \end{tabular}
}
\label{tab:ablation_edge}
\end{table}

\begin{table}
  \caption{Ablation study on dwell branch in STAR in terms of JSD. 
  All experimental results are conducted over five trials for a fair comparison.
  DB, MSC and SGP are short for the dwell branch, Moscow and Singapore. 
  A lower JSD value indicates a better performance.}
  \centering
  \resizebox{1.0\linewidth}{!}{
    \begin{tabular}{llcccccc}
    \toprule
    Dataset & Method & Distance & Radius & Duration & DailyLoc & G-rank & I-rank \\
    \midrule
    \multirow{3}{*}{NYC} & w/o DB & 0.1149 & 0.3638 & 0.0015 & 0.0615 & 0.0441 & 0.0598 \\
    & STAR & 0.1134 & 0.3615 & 0.0012 & 0.0597 & 0.0378 & 0.0550 \\
    & \% Improv. & 4.08\% & 2.18\% & 15.38\% & 0.34\% & 4.52\% & 5.31\% \\
    \midrule
    \multirow{3}{*}{TKY} & w/o DB & 0.1322 & 0.4484 & 0.0007 & 0.0379 & 0.0371 & 0.0644 \\
    & STAR & 0.1206 & 0.4198 & 0.0004 & 0.0340 & 0.0307 & 0.0650 \\
    & \% Improv. & 8.77\% & 6.37\% & 45.71\% & 10.40\% & 17.30\% & - \\
    \midrule
    \multirow{3}{*}{MSC} & w/o DB & 0.0155 & 0.0247 & 0.0041 & 0.0531 & 0.0266 & 0.0606 \\
    & STAR & 0.0071 & 0.0165 & 0.0022 & 0.0516 & 0.0196 & 0.0518 \\
    & \% Improv. & 54.19\% & 33.36\% & 45.37\% & 2.90\% & 26.17\% & 14.49\% \\
    \midrule
    \multirow{3}{*}{SGP} & w/o DB & 0.0123 & 0.0126 & 0.0018 & 0.0341 & 0.0439 & 0.0485 \\
    & STAR & 0.0080 & 0.0108 & 0.0009 & 0.0330 & 0.0283 & 0.0418 \\
    & \% Improv. & 34.96\% & 14.44\% & 48.89\% & 3.17\% & 35.44\% & 13.86\% \\
    \bottomrule
    \end{tabular}
  }
  \label{tab:ablation_stay}
\end{table}

\begin{figure*}
  \centering
  \subfloat{\includegraphics[width=0.3\textwidth]{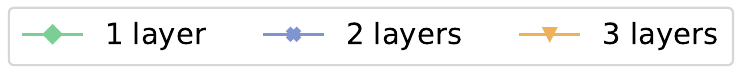}}\\
  \vspace{-1em}
  \subfloat{\includegraphics[width=\textwidth]{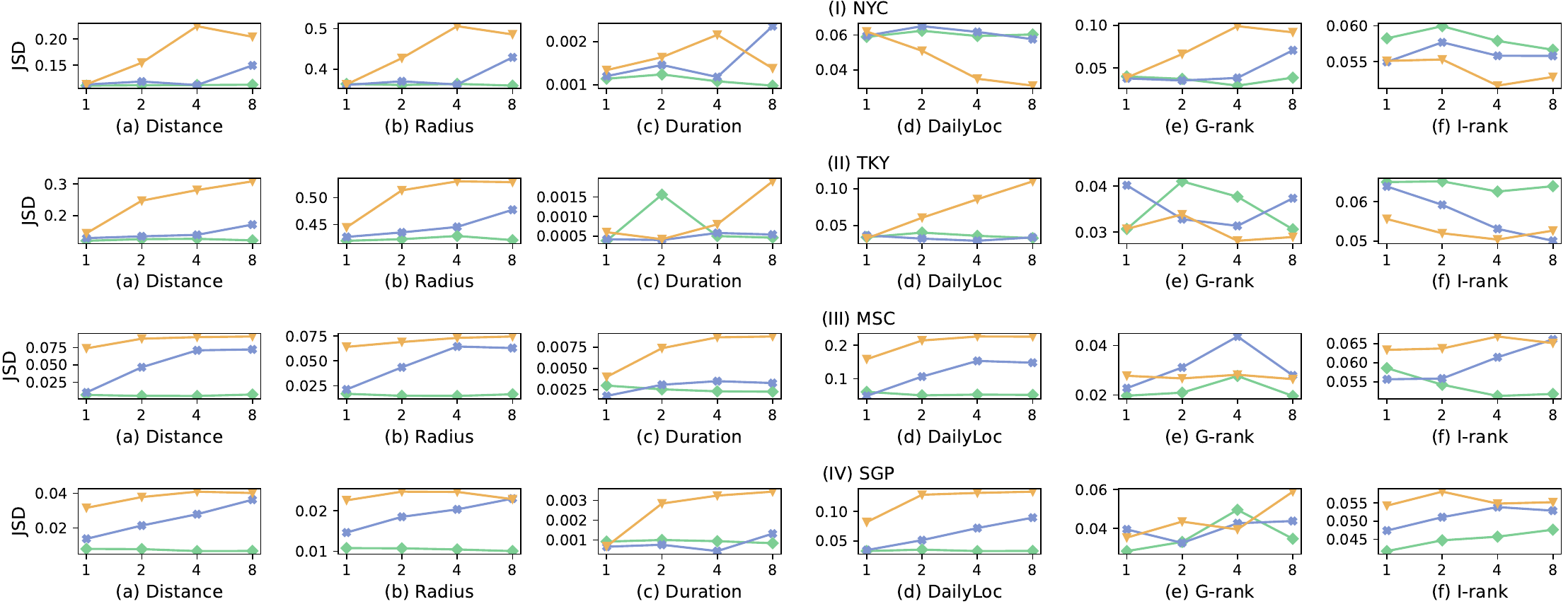}}
  \caption{Effects of the number of layers and attention heads on STAR. All experimental results are conducted over five trials.}
  \label{fig:layer_head}
\end{figure*}

In this part, we attempt to investigate the effectiveness of different modules in the STAR framework.
\subsubsection{Designs of Multi-channel Graph}
In order to learn the spatiotemporal correspondence among locations, we design three kinds of spatiotemporal graphs in the multi-channel embedding module to represent the node embeddings separately and obtain the final embeddings with a fusion layer.
To verify the effectiveness of the three graphs in STAR, we get rid of any of them (i.e., without SpatioTemporal Graph, without Spatial Distance Graph and without Temporal Transition Graph) to learn the node embeddings and perform human mobility simulation.
Additionally, we perform the experiment without any spatiotemporal graph to validate the effectiveness of the entire Multi-channel Embedding Module.
The comparison of their simulated results is shown in Figure~\ref{fig:gid_res}.

Obviously, the performance significantly deteriorates when spatiotemporal graphs are not used (i.e. w/o ALL), which suggests the necessity of incorporating spatiotemporal graphs for locations.
Compared with the results predicted by removing any of the three graphs, the fused embedding achieves optimal performance on almost all metrics over the four datasets.
In addition, the average performance of removing SpatioTemporal Graph is the worst on six metrics, especially on \textit{Distance}, \textit{Radius}, \textit{G-rank} and \textit{I-rank} metrics, which measure the effectiveness of simulated trajectories from both spatial and preference perspectives.
It indicates that relying solely on the static distance and transition information of locations is not enough to carry out accurate trajectory simulation.

Eliminating the Spatial Distance Graph results in a decline of performance on distance-based metrics (i.e. \textit{Distance} and \textit{Radius}), which is as per our hypothesis.
However, it scores well on temporal and preference metrics (i.e. \textit{Duration} and \textit{G-rank}), which is attributed to the SpatioTemporal Graph and Temporal Transition Graph effectively learning the temporal patterns and the frequently visited locations in the trajectories.
Furthermore, removing the Temporal Transition Graph results in inferior performance compared with remaining all three graphs but surpasses removing any other graph overall, which suggests that the SpatioTemporal Graph can aptly supplement the temporal transition information of trajectories.

\subsubsection{Designs of Edge Type in GNNs}
In order to explore the effect of edge weight in GNNs on the model performance, we set two edge types for model, \textit{Weighted} and \textit{Vanilla}. 
\textit{Weighted} retains the weights of edges in the three graphs, that is, the three graphs before binarization, while \textit{Vanilla} only uses the adjacency relationship of the three graphs to learn node embedding.
We list their performance in Table~\ref{tab:ablation_edge}.

It can be seen from the results that \textit{Vanilla} is better than \textit{Weighted} on almost all metrics of the four datasets, especially on the \textit{Distance} metric of the Moscow and Singapore dataset, the improvement reaches 31.07\% and 28.57\% , respectively. 
The reason for the observed result is that the original graph structure may not be optimal or informative~\cite{wu2022nodeformer, zhou2023opengsl}.
Firstly, \textit{Weighted} edges may introduce additional noise or bias during the learning process, leading  to suboptimal or inaccurate models.
Secondly, \textit{Vanilla} edges can simplify the learning process and yield a more robust and generalizable representation, thereby improving the performance on the human mobility simulation task.

The six metrics can be classified into three categories based on the level of improvement they show when comparing the \textit{Vanilla} graph to the \textit{Weighted} graph. 
Distance-based metrics (i.e. \textit{Distance} and \textit{Radius}) exhibit the highest improvement, whereas preference-based metrics (i.e. \textit{G-rank} and \textit{I-rank}) show the least improvement.
Time-dependent metrics (i.e. \textit{Duration} and \textit{Dailyloc}) showcase moderate improvements.
The underlying reason lies in the fact that our proposed multi-channel graphs explicitly encode the distance and temporal transitions among locations, while lacking user preference constraints.
Additionally, the long-tailed distribution of the \textit{Weighted} version might exacerbate the uneven distribution of location frequencies.

\subsubsection{Designs of Dwell Branch}
The dwell branch which allows individuals to remain in the current location with a learnable probability, aims to adaptively perceive the varying durations in mobility trajectories. 
To verify its effectiveness, we conduct the ablation experiments shown in Table~\ref{tab:ablation_stay}.
As we can observe, the design of the dwell branch enhances the performance on twenty three metrics over twenty four metrics of the four datasets, with the most significant improvement observed with the \textit{Distance} metric on the Moscow dataset, which is up to 54.19\%.
It implies that enabling individuals to dwell in a specific location with a learnable probability can facilitate a more comprehensive representation of the complex real-world behavior, consequently improving the performance of various metrics from different perspectives.

Specifically, there are two reasons for the performance improvement.
First of all, staying at the current location with a certain probability reduces the likelihood of long-distance movement for an individual. 
As a result, the dwell branch facilitates the acquisition of distance-related patterns and features, subsequently improving the performance of distance-based and time-dependent metrics.
Secondly, individuals tend to frequently visit several specific locations and thereby the learnable stay probability can increase individuals' visit probability at frequently visited places, which effectively captures the mobility preferences of real trajectories from both collective and individual views.
It is worth noting that the dwell branch achieves the most significant performance improvement on the Moscow dataset, as expected.
Due to the high frequency of staying in the previous location in the Moscow dataset, the learnable stay probability is essential for modelling repeated movement patterns within the trajectories more effectively.

\subsection{RQ3: Parameter Sensitivity of STAR}
\label{sec:param}
Parameter sensitivity analysis can help us understand how changes in parameters of model affect its performance, which is conducive to optimizing the model's performance, identifying key parameters that have the most important impact on results, and assessing the reliability and robustness of the model.
We plot the grid search results of the number of GAT layers over \{1, 2, 3\} and the number of attention heads over \{1, 2, 4, 8\} to thoroughly test the STAR framework on twenty four metrics over the four datasets. 
As can be seen from Figure~\ref{fig:layer_head}, the performance of STAR is robust under different hyperparameter settings, which indicates different hyperparameter values wouldn't affect its superiority over other baselines.

Specifically, the performance of different hyperparameters varies from metric to metric.
On distance-based metrics (i.e., \textit{Distance} and \textit{Radius}), only a simple model with a few number of GAT layers and attention heads can achieve good performance.
As the number of layers and attention heads increases, the performance of the model on distance-based metrics deteriorates due to overfitting.
The phenomenon emerges since moving distance is the most conspicuous pattern of trajectories, where the distance-related features in trajectories can be effectively modeled with fewer GAT layers and attention heads. 
On time-dependent metrics (i.e., \textit{Duration} and \textit{Dailyloc}), the performance of the model exhibits minimal fluctuations with the changes in the number of GAT layers and attention heads, which demonstrates the robustness of our proposed method.
On preference-based metrics (i.e., \textit{G-rank} and \textit{I-rank}), a simple model with a few number of GAT layers and attention heads performs poorly. 
Firstly, deeper GAT layers improves the performance of the model, which is because deeper layers can effectively capture complex and advanced graph structures.
Secondly, the increase in the number of attention heads also improves the model's performance.
Multiple attention heads enable the model to capture different and potentially complementary information from the neighborhood of each node, improving the quality of the learned node representation.

\begin{figure}[t]
  \centering
  \includegraphics[width=\linewidth]{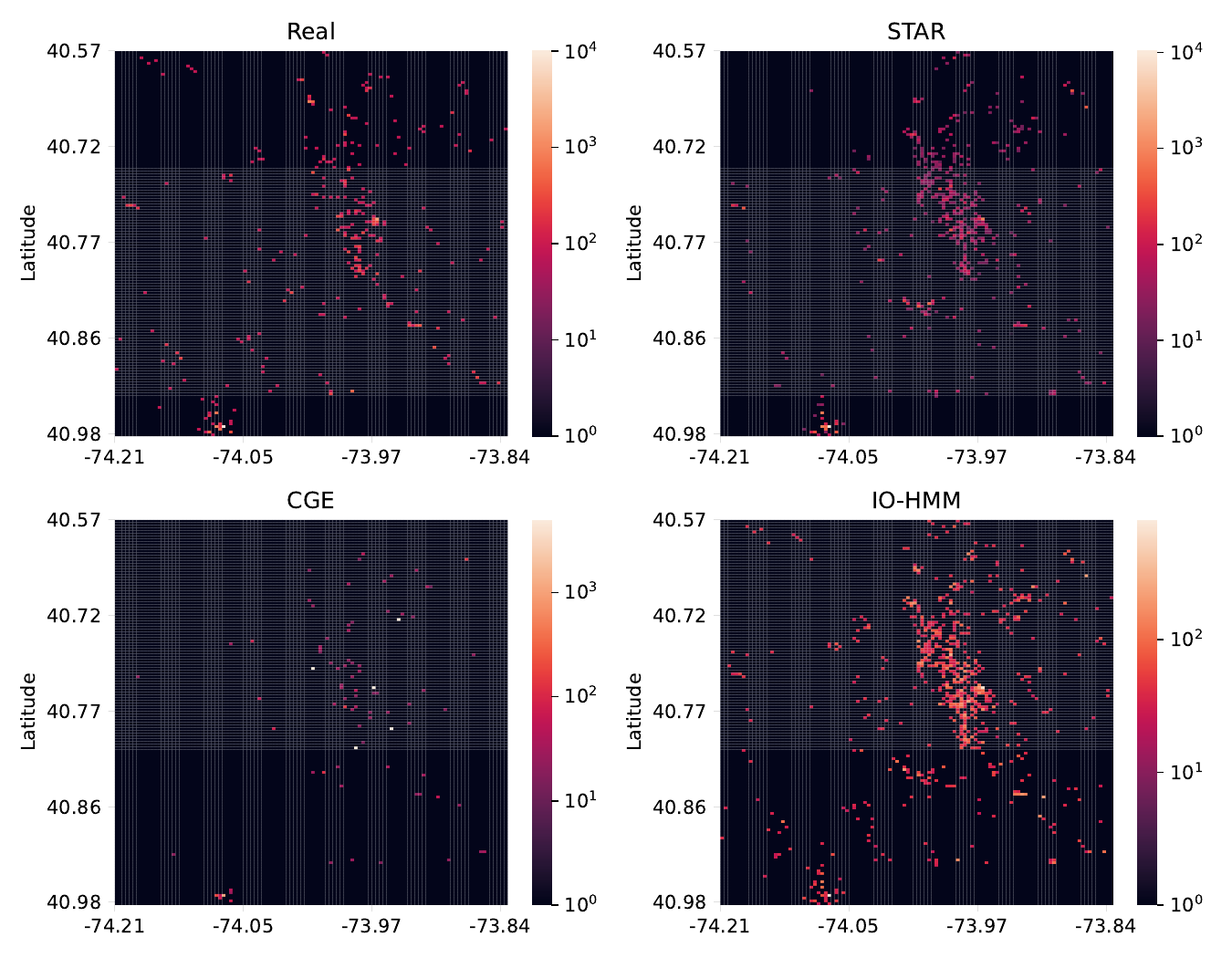}
  \caption{The geographical visualization of location visit frequency of real and simulated human trajectories on the NYC dataset.}
  \label{fig:NYC-heatmap}
\end{figure}

\subsection{RQ4: Visualization}
We visualize the location visit frequency of all methods on the four datasets to distinguish their simulated trajectories from the preference perspective (i.e. \textit{G-rank} metric).
Due to space limitations, we present the results of the real data and three representative methods on the first dataset (NYC) in Figure~\ref{fig:NYC-heatmap}.
It can be observed that the location visit frequency scale of STAR is closer to the real data compared with the other two baseline methods.
However, the visited locations in trajectories simulated by CGE are more concentrated, where some highly visited locations are even not visited in the real data.
Conversely, IO-HMM exhibits more attentions on the lower-visited areas, resulting in a relatively scattered distribution of visited locations.

\section{Conclusion}
In this paper, we propose a novel framework for human mobility simulation, namely \textbf{S}patio\textbf{T}emporal-\textbf{A}ugmented g\textbf{R}aph neural networks~(STAR), to model the spatiotemporal effects among locations.
On the one hand, STAR designs various kinds of spatiotemporal graphs to incorporate the spatiotemporal correspondences of locations, revealing the spatial proximity and functional similarity with the temporal visit distribution.
On the other hand, STAR builds the dual-branch decision generator module to balance the diverse transitions generated by the exploration branch and the repetitive patterns generated by the dwell branch.
The STAR framework is optimized based on the classification rewards of the policy discriminator module after iteratively generating a complete trajectory sequence.
We have conducted comprehensive experiments on the human mobility simulation task, verifying the superiority of the STAR framework to the \textit{state-of-the-art} methods.
Ablation studies on the multi-channel embedding module reveal that different datasets prefer different spatiotemporal graphs.
Elaborate experiments further verify the effectiveness of the edges of the \emph{Vanilla} version and the dwell branch.

\textbf{Limitations and future works}.  
The limitations of our work stem from the lack of dynamic scenarios and data scarcity.
Firstly, our proposed STAR lacks consideration for dynamic scenarios, such as massive commuting patterns and traffic accidents, which may result in an inaccurate capture of real-world human mobility behaviors and limit its applicability.
Secondly, similar to other deep learning methods, our proposed STAR may encounter performance degradation in scenarios with limited data availability for human mobility simulation. 
In future work, we plan to consider more external factors as well as mobility commonalities of different human mobility data to comprehensively enhance the mobility simulation task.

\section{Ackowledgement}
This research is supported by the Joint Funds of the Zhejiang Provincial Natural Science Foundation of China under (Grant No. LHZSD24F020001), Zhejiang Province "LingYan" Research and Development Plan Project (No. 2024C01114).

\ifCLASSOPTIONcaptionsoff
  \newpage
\fi

\bibliographystyle{IEEEtran}
\bibliography{ref}

\end{document}